\pdfoutput=1

\documentclass[11pt]{article}

\usepackage[]{acl}

\usepackage{times}
\usepackage{latexsym}

\usepackage[T1]{fontenc}

\usepackage[utf8]{inputenc}

\usepackage{microtype}

\usepackage{graphicx}
\usepackage{amsmath}
\usepackage{amsthm}
\usepackage{booktabs}
\usepackage{color}
\usepackage{xcolor}
\usepackage{xspace}
\usepackage{subcaption}
\usepackage{makecell}
\usepackage{fontawesome}
\usepackage[export]{adjustbox}
\usepackage{tabularx}
    \newcolumntype{L}{>{\raggedright\arraybackslash}X}

\definecolor{twitter}{RGB}{29, 161, 242}
\definecolor{reddit}{RGB}{255, 67, 1}
\definecolor{newspaper}{RGB}{24, 44, 97}

\newcommand\twitterLogo{\textcolor{twitter}{\faTwitter}\xspace}
\newcommand\redditLogo{\textcolor{reddit}{\faReddit}\xspace}
\newcommand\newspaperLogo{\textcolor{newspaper}{\faNewspaperO}\xspace}

\newcommand\papertitle{A Survey on Stance Detection for Mis- and Disinformation Identification}

\title{\papertitle}

\author{
Momchil Hardalov$^{1,2}$ \quad
Arnav Arora$^{1,3}$ \quad
Preslav Nakov$^{1,4}$ \quad
Isabelle Augenstein$^{1,3}$ \\
$^1$Checkstep Research\\
$^2$Sofia University ``St. Kliment Ohridski'', Bulgaria\\
$^3$University of Copenhagen, Denmark\\
$^4$Qatar Computing Research Institute, HBKU, Doha, Qatar\\
{\tt \{momchil, arnav, preslav.nakov, isabelle\}@checkstep.com}
}

\begin{document}
\maketitle

\begin{abstract}
Understanding attitudes expressed in texts, also known as \textit{stance detection}, plays an important role in systems for detecting false information online, be it misinformation (unintentionally false) or disinformation (intentionally false information). Stance detection has been framed in different ways, including (a)~as a component of fact-checking, rumour detection, and detecting previously fact-checked claims, or (b)~as a task in its own right. While there have been prior efforts to contrast stance detection with other related tasks such as argumentation mining and sentiment analysis, there is no existing survey on examining the relationship between stance detection and mis- and disinformation detection. Here, we aim to bridge this gap by reviewing and analysing existing work in this area, with mis- and disinformation in focus, and discussing lessons learnt and future challenges.
\end{abstract}

\section{Introduction}
 
The past decade is characterized by a rapid growth in popularity of social media platforms
such as Facebook, Twitter, Reddit, and more recently, Parler.
This, in turn, has led to a flood of dubious content, especially during controversial events such as 
Brexit and the US presidential election.
More recently, with the emergence of the COVID-19 pandemic, social media were at the center of the first global infodemic~\cite{alam2020fighting}, thus raising yet another red flag and a reminder of the 
need for effective mis- and disinformation detection online.

In this survey, we examine the relationship between automatically detecting false information online -- including fact-checking, and detecting fake news, rumors, and hoaxes -- and the core underlying Natural Language Processing (NLP) task needed to achieve this, namely \emph{stance detection}. 
Therein, we consider mis- and disinformation, which both refer to false information, though disinformation has an additional intention to harm.

\begin{table*}[t]
    \centering
    \resizebox{\textwidth}{!}{%
    \setlength{\tabcolsep}{4pt}
    \begin{tabular}{l|cllcll}
        \toprule
        \bf{Dataset}  & \bf{Source(s)}  & \bf{Target} & \bf{Context}  & \bf{Evidence}  & \bf{\#Instances}  & \bf{Task}  \\
         \midrule
         \multicolumn{7}{c}{\bf{English Datasets}} \\
         \textit{Rumour Has It}~{\cite{qazvinian-etal-2011-rumor}} 
         & \twitterLogo & Topic & Tweet & \faTh & 10K & Rumours \\
         \textit{PHEME}~{\cite{zubiaga-etal-2016-rumor-spread}} & \twitterLogo & Claim & Tweet & \faCommentsO & 4.5K & Rumours \\
         \textit{Emergent}~{\cite{ferreira-vlachos-2016-emergent}} & \newspaperLogo & Headline & Article$^*$ & \faTh & 2.6K & Rumours  \\
         \textit{FNC-1}~{\cite{pomerleau-2017-FNC}} & \newspaperLogo & Headline & Article & \faFileTextO & 75K & Fake news  \\
         \textit{RumourEval~'17}~{\cite{derczynski-etal-2017-rumoureval}} & \twitterLogo & Implicit\footnotemark[1] & Tweet & \faCommentsO & 7.1K & Rumours \\ 
         \textit{FEVER}~{\cite{thorne-etal-2018-fever}} & \faWikipediaW & Claim & Facts & \faTh & 185K & Fact-checking \\
         \textit{Snopes}~{\cite{hanselowski-etal-2019-snopes}} & Snopes & Claim & Snippets & \faTh & 19.5K & Fact-checking \\
         \textit{RumourEval~'19}~{\cite{gorrell-etal-2019-semeval}} & \twitterLogo~\redditLogo & Implicit\footnotemark[1] & Post & \faCommentsO & 8.5K & Rumours \\ 
         \textit{COVIDLies}~{\cite{hossain-etal-2020-covidlies}} & \twitterLogo & Claim & Tweet & \faFileTextO & 6.8K & Misconceptions \\
         \textit{TabFact}~{\cite{Chen2020TabFact}} & \faWikipediaW & Statement & WikiTable & \faTh & 118K & Fact-checking \\
         \midrule
         \multicolumn{7}{c}{\bf{Non-English Datasets}} \\
         \textit{Arabic FC}~{\cite{Baly-2018-IntegratingSD}} & \newspaperLogo & Claim & Document & \faFileTextO & 3K & Fact-checking \\
         \textit{DAST (Danish)}~{\cite{lillie-etal-2019-joint}} & \redditLogo & Submission & Comment & \faCommentsO & 3K & Rumour \\
         \textit{Croatian}~{\cite{bosnjak-karan-2019-data}} & \newspaperLogo & Title & Comment & \faFileTextO & 0.9K & Claim verifiability \\
         \textit{ANS (Arabic)}~{\cite{khouja-2020-stance}} & \newspaperLogo & Claim & Title & \faFileTextO & 3.8K & Claim verification \\
         \textit{Ara(bic)Stance}~{{\citep{AraStance2021:NLP4IF}}} & \newspaperLogo & Claim & Title & \faFileTextO & 4K & Claim verification \\
         \bottomrule
    \end{tabular}
    }
    \caption{Key characteristics of stance detection datasets for mis- and disinformation detection. \emph{\#Instances} denotes dataset size as a whole; the numbers are in thousands (K) and are rounded to the hundreds. $^*$the article's body is summarised. 
    \textit{Sources}: \twitterLogo~Twitter, \newspaperLogo~News, {\faWikipediaW}ikipedia, \redditLogo~Reddit. \textit{Evidence}: \faFileTextO~Single, \faTh~Multiple, \faCommentsO~Thread.}
    \label{tab:dataset_features}
\end{table*}

Detecting and aggregating the expressed stances towards a piece of information can be a powerful tool for a variety of tasks including understanding ideological debates~\cite{Hasan2014WhyAY}, gathering different frames of a particular issue~\cite{shurafa-etal-2020-stance-framing} or determining the leanings of 
media outlets~\cite{stefanov-etal-2020-predicting}. The task of stance detection has been studied from different angles, e.g.,~in political debates~\cite{habernal-etal-2018-arc}, for fact-checking~\cite{thorne-etal-2018-fever}, or regarding new products
~(\href{{https://}}{Somasundaran et al., 2009}).
Moreover, different types of text have been studied, including social media posts~\cite{zubiaga-etal-2016-rumor-spread} and news articles~\cite{pomerleau-2017-FNC}. Finally, stances expressed by different actors have been considered, such as politicians
~(\href{{https://}}{Johnson et al., 2009}),
journalists~\cite{hanselowski-etal-2019-snopes}, and users on the web~\cite{derczynski-etal-2017-rumoureval}. 

There are some recent surveys related to stance detection.
\citet{zubiaga-et-al-2018-socialmedia-survey} discuss the role of stance in rumour verification, %
\citet{ALDAYEL2021102597} survey stance detection for social media, %
and \citet{kucuk-2020-stance-survey} 
survey stance detection holistically, without a specific focus on veracity.
There are also surveys on fact-checking~\cite{thorne-vlachos-2018-automated,10.1162/tacl_a_00454}, which mention, though do not exhaustively survey, stance.

However, there is no existing overview of the role that different formulations of stance detection play in the detection of false content. In that respect, stance detection could be modelled as fact-checking --- to gather the stances of users or texts towards a claim or a headline (and support fact-checking or studying misinformation) ---, or as a component of a system that uses stance as part of its process of judging the veracity of an input claim. Here, we aim to bridge this gap by surveying the research on stance for mis- and disinformation detection, including task formulations, 
datasets, and methods, from which we draw conclusions and lessons, and we forecast future research trends.

\section{What is Stance?}
\label{sec:stance}

In order to understand the task of stance detection, we first provide definitions of stance and the stance-taking process.  
\citet{doi:10.1080/01638538809544689} define stance as the expression of a speaker's standpoint and judgement towards a given proposition. Further, \citet{du2007stance}) define stance as ``\emph{a public act by a social actor, achieved dialogically through overt communicative means, of simultaneously evaluating objects, positioning subjects (self and others), and aligning with other subjects, with respect to any salient dimension of the sociocultural field}'', showing that the stance-taking process is affected not only by personal opinions, but also by other external factors such as cultural norms, roles in the institution of the family, etc. 
Here, we adopt the general definition of stance detection by~\citet{kucuk-2020-stance-survey}: 
``\emph{for an input in the form of a piece of text and a target pair, stance detection is a classification problem where the stance of the author of the text is sought in the form of a category label from this set: {Favor, Against, Neither}. Occasionally, the category label of Neutral is also added to the set of stance categories~\cite{mohammad-etal-2016-semeval}, and the target may or may not be explicitly mentioned in the text}''~\cite{augenstein-etal-2016-stance,mohammad-etal-2016-semeval}. Note that the stance detection definitions and the label inventories vary somewhat, depending on the target application (see Section~\ref{sec:factuality}).

Finally, stance detection can be distinguished from several other closely related NLP tasks: 
(\emph{i})~\emph{biased language detection}, where the existence of an inclination or tendency towards a particular perspective within a text is explored,
(\emph{ii})~\emph{emotion recognition}, where the goal is to recognise emotions such as \emph{love, anger, etc.} in the text,
(\emph{iii})~\emph{perspective identification}, which aims to find the point-of-view of the author (e.g.,~Democrat vs. Republican) and the target is always explicit,
(\emph{iv})~\emph{sarcasm detection}, where the interest is in satirical or ironic pieces of text, often written with the intent of ridicule or mockery, and
(\emph{v})~\emph{sentiment analysis}, which checks the polarity of a piece of text.

\section{Stance and Factuality}
\label{sec:factuality}

\begin{figure*}[t]
    \centering
    \subfloat[stance detection as fact-checking \label{fig:formulation:fc}]{
        \includegraphics[width=0.9\columnwidth,valign=t]{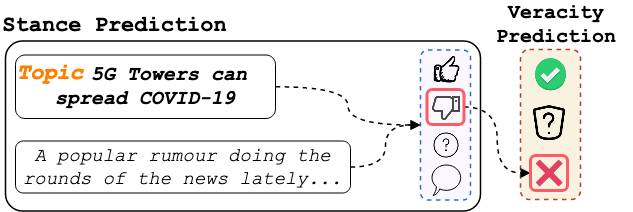}
    }
    \subfloat[stance detection as a component of a fact-checking pipeline \label{fig:formulation:comp}]{
    \includegraphics[width=1.1\columnwidth,valign=t]{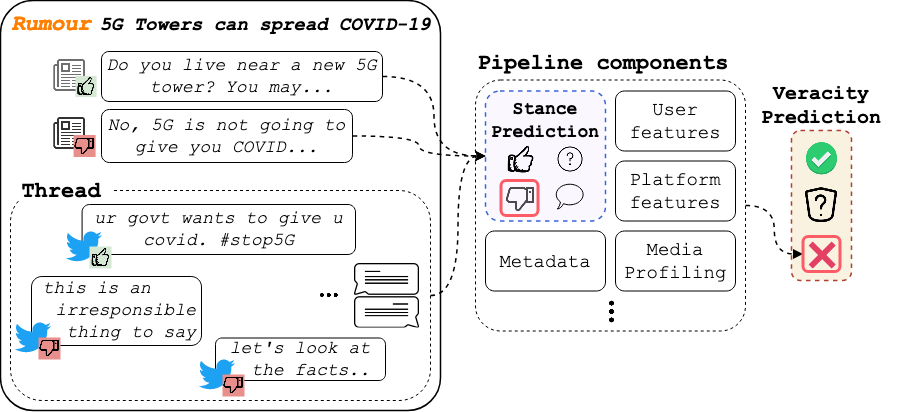}
    }
    \caption{Two stance detection formulations.}
    \label{fig:formulation}
\end{figure*}

\begin{figure}[t]
    \centering
    \includegraphics[width=\columnwidth]{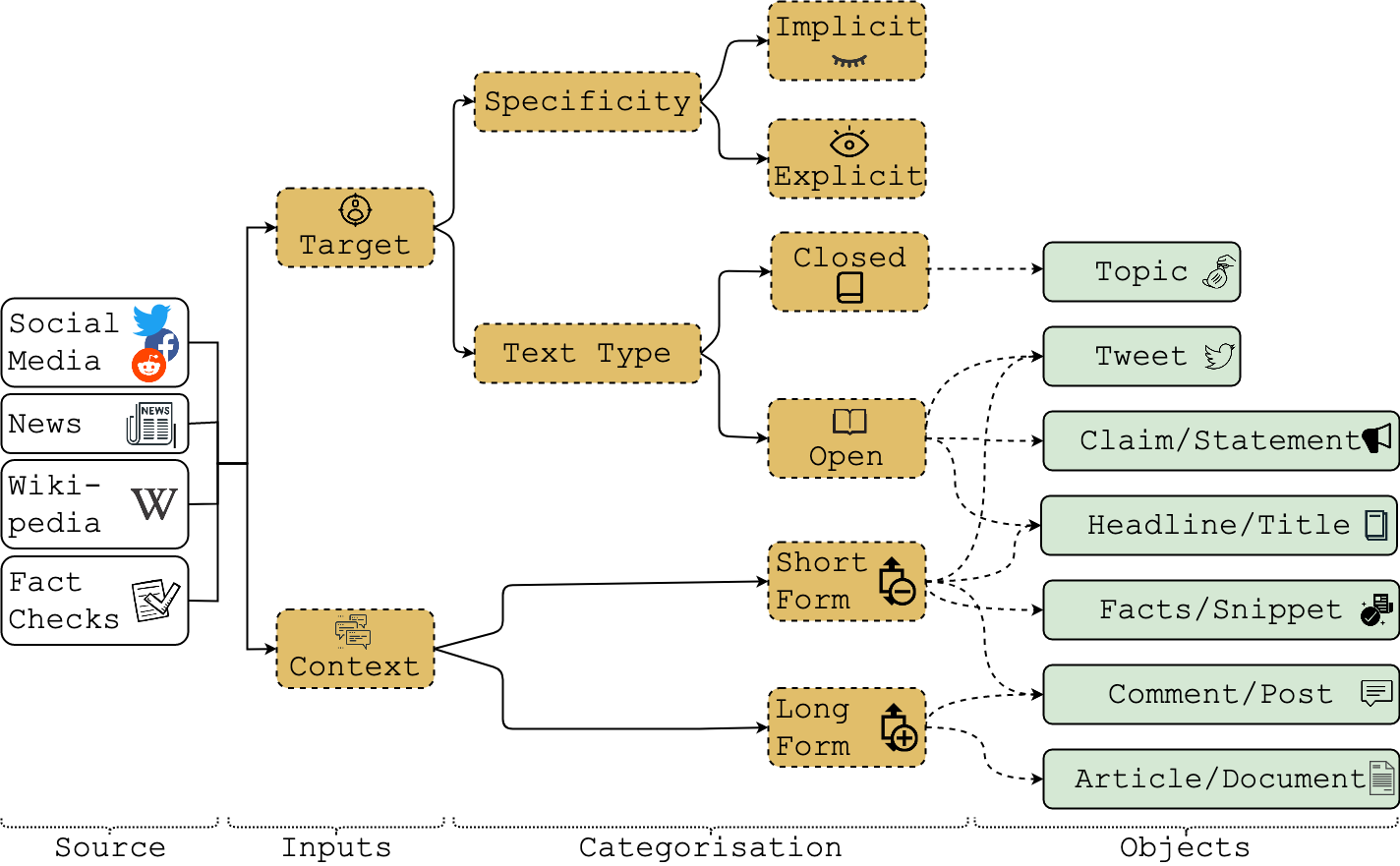}
    \caption{
    Types of stance. The \emph{Target} is the object of the stance expressed in the \emph{Context}. 
    }
    \label{fig:tree}
\end{figure}

Here, we offer an overview of the settings for mis- and disinformation identification to which stance detection has been successfully applied.
As shown in Figure~\ref{fig:formulation}, stance can be used (a)~as a way to perform fact-checking, or more typically, (b)~as a component of a fact-checking pipeline.
Table~\ref{tab:dataset_features} 
shows an overview of the key characteristics of the available datasets.  %
We include the \emph{source} of the data and the \emph{target}\footnote{The target can either be explicit, e.g.,~a topic such as \emph{Public Healthcare}, or implicit, where only the context is present and the target is not directly available and is usually a topic~\citep{derczynski-etal-2017-rumoureval,gorrell-etal-2019-semeval}, e.g., \emph{Germanwings}, or `\emph{Prince to play in Toronto}'. When the target is implicit, the task becomes similar to sentiment analysis.} towards which the stance is expressed in the provided textual \emph{context}. 
\newline
We further show the type of evidence: \emph{Single} is a single document/fact, \emph{Multiple} is multiple pieces of textual evidence, often facts or documents, \emph{Thread} is a (conversational) sequence of posts or a discussion. The final column is the type of the target \emph{Task}. 
{Finally, we present a dataset-agnostic summary of the terminology used for the different types of stance (see Figure~\ref{fig:tree}), which we describe in a four-level taxonomy: (\emph{i})~sources, i.e.,~where the dataset was collected from, (\emph{ii})~inputs that represent the stance target (e.g.,~claim), and the accompanying context (e.g.,~news article), (\emph{iii})~categorisation -- meta-level characteristics of the input, and (\emph{iv})~the textual object types for a particular stance scenario (e.g.,~topic, tweet, etc.).}
Appendix~\ref{sec:appx:examples} discusses different stance scenarios with corresponding contexts and targets, with illustrations in Table~\ref{tab:examples}.

\subsection{Fact-Checking as Stance Detection}
\label{sec:factuality:fc_as_stance}

As stance detection is the core task within fact-checking, prior work has studied it in isolation, e.g.,~predicting the stance towards one or more documents. {More precisely, the stance of the textual evidence(s) toward the target claim is considered as a veracity label, as illustrated in Figure~\ref{fig:formulation:fc}.}

\paragraph{Fact-Checking with One Evidence Document} 
\citet{pomerleau-2017-FNC} organised the first Fake News Challenge
(FNC-1) with the aim of automatically detecting fake news.  %
The goal was to detect the relatedness of a news article's body w.r.t. a headline (possibly from another news article), based on the stance that the former takes regarding the latter. The possible categories were \emph{positive}, \emph{negative}, \emph{discuss}, and \emph{unrelated}.
This was a standalone task, as it provides stance annotations only, omitting the actual ``truth labels'', with the motivation of assisting fact-checkers in gathering several distinct arguments pertaining to a particular claim.%

\paragraph{Fact-Checking with Multiple Evidence Documents} The FEVER~\cite{thorne-etal-2018-fever,thorne-etal-2019-fever2} shared task was introduced in 2018, aiming to determine the veracity of a claim based on a set of statements from Wikipedia. Claims can be composite and can contain multiple (contradicting) statements, which requires multi-hop reasoning, and the claim--evidence pairs are annotated as \emph{SUPPORTED}, \emph{REFUTED}, and \emph{NOT ENOUGH INFO}. The latter category includes claims that are either too general or too specific, and cannot be supported or refuted by the available information in Wikipedia. This setup may help fact-checkers understand the decisions a model made in their assessment of the veracity of a claim, or assist human fact-checkers.

The second edition (\citeyear{thorne-etal-2019-fever2}) of FEVER evaluated the robustness of models to adversarial attacks, where the participants were asked to provide new examples to ``break'' existing models, then to propose ``fixes'' for the system against such attacks. 

{Note that FEVER 
slightly differs from typical stance detection, as it considers evidence supporting or refuting a claim, rather than the stance of an author 
towards a claim. An alternative way to look at this is in terms of argument reasoning, i.e.,~extracting and providing factual evidence for a claim. 

FEVER also has a connection to Natural Language Inference, i.e.,~determining the relationship between two sentences. We view FEVER as requiring stance detection as it resembles FNC, which is commonly seen as a stance detection task.}

Apart from FEVER, \citet{hanselowski-etal-2019-snopes} presented a task constructed from manually fact-checked claims on Snopes.
For this task, a model had to predict the stance of evidence sentences in articles written by journalists towards claims. Unlike FEVER, this task does not require multi-hop reasoning.

\citet{Chen2020TabFact} studied the verification of claims using tabular data. The TabFact dataset was generated by human annotators who created positive and negative statements about Wikipedia tables. Two different forms of reasoning in a statement are required: (\emph{i})~linguistic, i.e.,~semantic understanding, and (\emph{ii}) symbolic, i.e.,~using the table structure.

\subsection{Stance as a (Mis-/Dis-)information Detection Component}
\label{sec:factuality:component}

Fully automated systems can assist in gauging the extent and studying the spread of false information online. 
This is in contrast to the previously discussed applications of stance detection -- as a stand-alone system for detecting mis- and disinformation.
Here, we review its potency to serve as a component in an automated pipeline. 
Figure~\ref{fig:formulation:comp} 
illustrates the setup, which can also include steps such as modelling the user or profiling the media outlet among others.
We discuss in more detail media profiling and misconceptions in Appendix~\ref{sec:appx:other_methods}.

\paragraph{Rumors}
Stance detection can be used for rumour detection and debunking, where the stance of the crowd, media, or other sources towards a claim are used to determine the veracity of a currently circulating story or report of uncertain or doubtful factuality. More formally, \emph{for a textual input and a rumour expressed as text, stance detection here is to determine the position of the text towards the rumour as a category label from the set \{Support, Deny, Query, Comment\}.} {\citet{zubiaga-etal-2016-rumor-spread} define these categories as whether the author: supports (\emph{Support}) or denies (\emph{Deny}) the veracity of the rumour they are responding to, ``asks for additional evidence in relation to the veracity of the rumour'' (\emph{Query}) or ``makes their own comment without a clear contribution to assessing the veracity of the rumour'' (\emph{Comment}). %
}
This setup was widely explored for microblogs and social media.

\citet{qazvinian-etal-2011-rumor} started with five rumours and classified the user's stance as \emph{endorse}, \emph{deny}, \emph{unrelated}, \emph{question}, or \emph{neutral}. While they were among the first to demonstrate the feasibility of this task formulation, the limited size of their study and the focus on assessing the stance of individual posts limited its real-world applicability. %

\citet{zubiaga-etal-2016-rumor-spread} analysed how people spread rumours on social media based on conversational threads. They included rumour threads associated with nine newsworthy events, and users' stance before and after the rumours were confirmed or denied. \citet{dungs-etal-2018-rumour} continued this line of research, but focused on the effectiveness of stance for predicting rumour veracity. \citet{hartmann-etal-2019-mapping} explored the flow of (dis-)information on Twitter after the MH17 Plane Crash. %

The two RumourEval~\cite{derczynski-etal-2017-rumoureval,gorrell-etal-2019-semeval} shared tasks on automated claim validation aimed to identify and handle rumours based on user reactions and ensuing conversations in social media, offering annotations for both stance and veracity. The two editions of RumourEval were similar in spirit, with the second one providing more tweets and also additionally Reddit posts. RumourEval demonstrated the importance of modelling the context of a story instead of drawing conclusions based on a single post.

\citet{ferreira-vlachos-2016-emergent} collected claims and news articles from rumour sites with annotations for stance and veracity by journalists as part of the Emergent
project. The goal was to use the stance of a news article, summarised into a single sentence, towards a claim as one of the components to determine its veracity. A downside is the need to summarise, in contrast to FNC-1~\cite{pomerleau-2017-FNC}, where entire news articles were used.

\paragraph{Multiple languages}
While the above research has focused exclusively or primarily on English, interest in stance detection for other languages has started to emerge. \citet{Baly-2018-IntegratingSD} integrated stance detection and fact-checking for Arabic in a single corpus. \citet{khouja-2020-stance} proposed a dataset for Arabic following the FEVER setup.
\citet{AraStance2021:NLP4IF} introduced {AraStance}, a multi-country and multi-domain dataset of Arabic stance detection for fact-checking.
\citet{lillie-etal-2019-joint} collected data for stance and veracity from Danish Reddit threads
\citet{zubiaga-etal-2016-rumor-spread}. \citet{bosnjak-karan-2019-data} studied stance detection and claim verification
of comments for Croatian news articles.

\section{Approaches}
\label{sec:methods}
In this section, we discuss various ways to use stance detection for mis- and disinformation detection,  and list the state-of-the-art results in Table~\ref{tab:sota}. %

\paragraph{Fact-Checking as Stance Detection}
Here, we discuss approaches for stance detection in the context of mis- and disinformation detection, where veracity is modelled as stance detection as outlined in Section~\ref{sec:factuality:fc_as_stance}. One such line of research is the Fake News Challenge,
which used weighted accuracy as an evaluation measure (FNC score), 
to mitigate the impact of class imbalance. Subsequently, \citet{Retrospective-C18-1158} criticized the FNC score and F1-micro, and argued in favour of F1-macro (F1)
instead. In the competition, most teams used hand-crafted features such as words, word embeddings, and sentiment lexica~\cite{riedel2017simple,Retrospective-C18-1158}. 
\citet{Retrospective-C18-1158} showed that the most important group of features were the lexical ones, followed by features from topic models, while sentiment analysis did not help.
\citet{ghanem-etal-2018-stance} investigated the importance of lexical cues,
and found that \emph{report} and \emph{negation} are most beneficial, while \emph{knowledge} and \emph{denial} are least useful. All these models struggle to learn the \emph{Disagree} class, achieving up to 18 F1 due to major class imbalance. In contrast, \emph{Unrelated} is detected almost perfectly by all models (over 99 F1). 
\citet{Retrospective-C18-1158} showed that these models exploit the lexical overlap between the headline and the document, but fail when there is a need to model semantic relations or complex negation, or to understand propositional content in general. This can be attributed to the use of $n$-grams, topic models, and lexica.

\citet{mitra2018memory} investigated memory networks, aiming to mitigate the impact of irrelevant and noisy information by learning a similarity matrix and a stance filtering component, %
and taking a step towards explaining the stance of a given claim by extracting meaningful snippets from evidence documents. Like previous work, their model performs poorly on the \emph{Agree/Disagree} classes, due to the unsupervised way of training the memory networks, i.e.,~there are no gold snippets justifying the document's stance w.r.t. the target claim. %

More recently, transfer learning with pre-trained Transformers has been explored~\citep{slovikovskaya-attardi-2020-transfer}, significantly improving the performance of previous state-of-the-art approaches.
\citet{guderlei-assenmacher-2020-evaluating} showed the most important hyper-parameter to be learning rate, while freezing layers did not help. In particular, using the pre-trained Transformer RoBERTa improved F1 from 18 to 58 for \emph{Disagree},  and from 50 to 70 for \emph{Agree}. The success of these models is also seen in cross-lingual settings. For Arabic, \citet{khouja-2020-stance} achieved 76.7 F1 for stance detection on the ANS dataset using mBERT. Similarly, \citet{hardalov2022fewshot} applied pattern-exploiting training (PET) with sentiment pre-training in a cross-lingual setting showing sizeable improvements on 15 datasets.
\citet{AraStance2021:NLP4IF} showed that language-specific pre-training was pivotal,
outperforming the state of the art on AraStance (52~F1) and Arabic FC (78 F1).

{Some formulations include an extra step for evidence retrieval,
e.g.,~retrieving Wikipedia snippets for FEVER~{\citep{thorne-etal-2018-fever}}. To evaluate the whole fact-checking pipeline, they introduced the FEVER score -- the proportion of claims for which both correct evidence is returned and a correct label is predicted.} The top systems that participated in the FEVER competition~\citet{hanselowski-etal-2018-ukp, yoneda-etal-2018-ucl, Nie2019CombiningFE}
used LSTM-based models for natural language inference, e.g.,~enhanced sequential inference model (ESIM~\citet{chen-etal-2017-enhanced}). \citet{Nie2019CombiningFE} proposed a neural semantic matching network, which ranked first in the competition, achieving 64.2 FEVER score. They used 
page view frequency and WordNet features in addition to pre-trained contextualized embeddings~\citep{peters-etal-2018-deep}.

More recent approaches used bi-directional attention~\cite{li-etal-2018-end}, a GPT language model~\citep{malon-2018-team,yang-etal-2019-blcu}, and graph neural networks~\cite{zhou-etal-2019-gear,atanasov-etal-2019-predicting,liu-etal-2020-fine,wang2020robust,zhong-etal-2020-reasoning,weinzierl2021misinformation, si-etal-2021-topic}. 
{\citet{zhou-etal-2019-gear} showed that adding graph networks on top of BERT can improve performance, 
reaching 67.1 FEVER score. Yet, the retrieval model is also important, e.g.,~using the gold evidence set adds 1.4 points. 
\citet{liu-etal-2020-fine,zhong-etal-2020-reasoning} replaced the retrieval model with a BERT-based one, in addition to using an improved mechanism to propagate the information between nodes in the graph, boosting the score to 70.
Recently, \citet{ye-etal-2020-coreferential} 
experimented with a retriever that %
incorporates co-reference in distant-supervised pre-training, namely, CorefRoBERTa.
\citet{wang2020robust} added external knowledge to build a contextualized semantic graph, 
setting a new SOTA on Snopes.
\citet{si-etal-2021-topic} and \citet{ijcai2021-536} improved multi-hop reasoning using 
a model with eXtra Hop attention~\citep{Zhao2020Transformer-XH}, a capsule network aggregation layer, and LDA topic information}. \citet{atanasova2022insufficient} introduced the task of evidence sufficiency prediction to more reliably predict the \textit{NOT ENOUGH INFO} class.

Another notable idea is to use pre-trained language models as fact-checkers based on a masked language modelling objective~\cite{lee-etal-2020-language}, or to use the perplexity of the entire claim with respect to the target document~\cite{lee-etal-2021-towards}. 
{Such models do not require a retrieval step, as they use the knowledge stored in language models. However, they are prone to biases in the patterns used, e.g.,~they can predict date instead of city/country and vice-versa when using ``born in/on''.
Moreover, the insufficient context can seriously confuse them, e.g.,~for short claims with uncommon words such as ``Sarawak is a ...'', where it is hard to detect the entity type. Finally, the performance of such models remains well below supervised approaches; even though recent work shows that \emph{few-shot training} can improve results}~\cite{lee-etal-2021-towards}.

{Error analysis suggests that the main challenges are (\emph{i}) confusing semantics at the sentence level, 
e.g.,~``\emph{Andrea Pirlo is an American professional footballer.}'' vs. ``\emph{Andrea Pirlo is an Italian professional footballer who plays for an American club.}'', 
(\emph{ii})~sensitivity to spelling errors, (\emph{iii})~lack of relation between the article and the entities in the claim, (\emph{vi})~dependence on syntactic overlaps,
 e.g.,~``\emph{Terry Crews played on the Los Angeles Chargers.}'' (\emph{NotEnoughInfo}) is classified as \emph{refuted}, given the sentence ``\emph{In football, Crews played ... for the Los Angeles Rams, San Diego Chargers and Washington Redskins, ...}'', 
(\emph{v})~embedding-level confusion, e.g.,~numbers tend to have similar embeddings, 
``\emph{The heart beats at a resting rate close to 22 bpm.}'' is not classified as \emph{refuted} based on the evidence sentence ``\emph{The heart beats at a resting rate close to 72 bpm.}'',
and similarly for months.
}

\paragraph{Threaded Stance}
{In the setting of conversational threads~\cite{zubiaga-etal-2016-rumor-spread,derczynski-etal-2017-rumoureval,gorrell-etal-2019-semeval}, in contrast to the single-task setup, which ignores or does not provide
further context, important knowledge can be gained from the structure of user interactions.}
{These approaches are mostly applied as part of a larger system, e.g.,~for detecting and debunking rumours (see Section~\ref{sec:factuality:component}, \emph{Rumours}).} A common pattern is to use tree-like structured models, fed with lexicon-based content formatting~\citep{zubiaga-etal-2016-stance} %
{or dictionary-based token scores~\citep{aker-etal-2017-simple}.} %
\citet{kumar-carley-2019-tree} replaced CRFs with Binarised Constituency Tree LSTMs, and used pre-trained embeddings to encode the tweets. More recently, Tree~\cite{ma-gao-2020-debunking} and Hierarchical~\cite{yu-etal-2020-coupled} Transformers were proposed, which combine post- and thread-level representations for rumour debunking, improving previous results on RumourEval~'17~\citep{yu-etal-2020-coupled}. %
\citet{kochkina-etal-2017-turing,kochkina-etal-2018-one} split conversations into branches, modelling each branch with branched-LSTM and hand-crafted features{, outperforming other systems at RumourEval~'17 on stance detection (43.4 F1)}. 
\citet{li-etal-2020-exploiting} deviated from this structure and modelled the conversations as a graph. \citet{lin-etal-2020-rumour-transfer} showed that pre-training on stance data yielded better representations for threaded tweets for downstream rumour detection. \citet{yang-etal-2019-blcu} 
{took a step further and}
curated per-class pre-training data by adapting examples, not only from stance datasets, but also from tasks such as question answering, achieving the highest F1 (57.9) on the RumourEval~'19 stance detection task. \citet{li-etal-2019-eventai,li-etal-2019-rumor} additionally incorporated
user credibility information, conversation structure, and other content-related features to predict the rumour veracity,
{ranking 3rd on stance detection and 1st on veracity classification (RumourEval~'19)}.
Finally, the stance of a post might not be expressed directly towards the root of the thread, thus the preceding posts must be also taken into account~\cite{gorrell-etal-2019-semeval}.

A major challenge for all rumour detection datasets is the class distribution~\citep{zubiaga-etal-2016-rumor-spread,derczynski-etal-2017-rumoureval,gorrell-etal-2019-semeval},
e.g., the minority class \emph{denying} is extremely hard for models to learn, as even for strong systems such as \citet{kochkina-etal-2017-turing} the F1 for it is 0. Label semantics also appears to play a role as the \emph{querying} label has a similar distribution, but much higher F1.
Yet another factor is thread depth, as performance drops significant at higher depth, especially for the \emph{supporting} class. On the positive side, using multi-task learning and incorporating stance detection labels into veracity detection yields a huge boost in performance~\cite{gorrell-etal-2019-semeval,yu-etal-2020-coupled}. 

Another factor, which goes hand in hand with the threaded structure, 
is the temporal dimension of posts in a thread~\citep{lukasik-etal-2016-hawkes,veyseh-etal-2017-temporal-attention,dungs-etal-2018-rumour,wei-etal-2019-modeling}.
{In-depth data analysis~(\citet{zubiaga-etal-2016-stance,zubiaga-etal-2016-rumor-spread,kochkina-etal-2017-turing,wei-etal-2019-modeling,ma-gao-2020-debunking,li-etal-2020-exploiting}; among others) shows interesting patterns along the temporal dimension: (\emph{i})~source tweets (at zero depth) usually support the rumour and models often learn to detect that, (\emph{ii})~it takes time for denying tweets to emerge, afterwards for false rumors their number increases quite substantially, (\emph{iii})~the proportion of querying tweets towards unverified rumors also shows an upward trend over time, but their overall number decreases}.

\paragraph{Multi-Dataset Learning (MDL)}
Mixing data from different domains and sources can improve robustness.
However, setups that combine mis- and disinformation identification with stance detection, outlined in Section~\ref{sec:factuality}, vary in their annotation and labelling schemes, which poses many challenges.

\footnotetext{{The result from \emph{dominiks} can be found at \url{https://competitions.codalab.org/competitions/18814\#results}}}

Earlier approaches focused on pre-training models on multiple tasks, e.g.,~\citet{fang-etal-2019-neural} achieved state-of-the-art results on FNC-1 by fine-tuning on multiple tasks such as question answering, natural language inference, etc., which are weakly related to stance. 
Recently, \citet{schiller2021stance} proposed a benchmark to evaluate the robustness of stance detection models.
They leveraged a pre-trained multi-task deep neural network, MT-DNN~\cite{liu-etal-2019-mt-dnn}, and continued its training on all datasets simultaneously using multi-task learning, showing sizeable improvements over models trained on individual datasets.
\citet{hardalov2021cross} experimented with cross-domain learning from 16 stance detection datasets. 
They proposed a novel
architecture (MoLE) that applies domain adaptation at different stages of the modelling process~\citep{1000251}: feature-level~\citep{guo-etal-2018-multi,wright-augenstein-2020-transformer} 
and decision-level~\citep{pmlr-v37-ganin15}.
They further integrated label embeddings~\citep{augenstein-etal-2018-multi}, and eventually developed an end-to-end unsupervised framework for predicting stance from a set of unseen target labels. %
\citet{hardalov2022fewshot} explored {PET}~\citep{schick-schutze-2021-exploiting} in a cross-lingual setting, %
combining datasets with different label inventories by modelling the task as a cloze question answering one. %

They showed that MDL helps for low-resource and substantively for full-resource scenarios. Moreover, transferring knowledge from English stance datasets and noisily generated sentiment-based stance data can further boost performance.

\paragraph{State of the Art}
\label{sec:appx:sota}
Table~\ref{tab:sota} shows the state-of-the-art (SOTA) results for each dataset discussed in Section~\ref{sec:factuality} and Table~\ref{tab:dataset_features}. The datasets vary in their task formulation and composition in terms of size, number of classes, class imbalance, topics, evaluation measures, etc. Each of these factors impacts the performance, leading to sizable differences in the final score, as discussed in Section~\ref{sec:methods}, and hence rendering the reported results hard to compare directly across these datasets.

\begin{table}[t]
    \centering
    \setlength{\tabcolsep}{3.2pt}
    \resizebox{1.00\columnwidth}{!}{%
    \begin{tabular}{llll}
    \toprule
        \bf{Paper} &  \bf{Dataset} & \bf{Score} & \bf{Metric} \\
        \midrule
         \citet{hardalov2021cross} & Rumour Has It & 71.2 & F1$_{macro}$ \\
        \href{http://}{Kumar et al.}~\shortcite{kumar-carley-2019-tree} & PHEME & 53.2 & F1$_{macro}$ \\
         \citet{hardalov2021cross} & Emergent & 86.2 & F1$_{macro}$ \\
         \href{http://}{Guderlei et al.}~\shortcite{guderlei-assenmacher-2020-evaluating} & FNC-1 & 78.2 & F1$_{macro}$ \\
         \citet{yu-etal-2020-coupled} & RumourEval~'17 & 50.9  & F1$_{macro}$  \\
         \href{https://competitions.codalab.org/competitions/18814\#results}{Dominiks (2021)}$^*$ & FEVER & 76.8 & FEVER \\
         \citet{wang2020robust} & Snopes & 78.3 & F1$_{macro}$  \\
         \citet{yang-etal-2019-blcu} & RumourEval~'19 & 61.9 & F1$_{macro}$ \\
         \citet{weinzierl2021misinformation} & COVIDLies & 74.3 & F1$_{macro}$  \\
         \citet{liu2022tapex} & TabFact & 84.2 & Accuracy   \\
         \midrule
         \citet{AraStance2021:NLP4IF} & Arabic FC & 52.? & F1$_{macro}$    \\
         \citet{lillie-etal-2019-joint} & DAST & 42.1 & F1$_{macro}$  \\
         \citet{bosnjak-karan-2019-data} & Croatian & 25.8 & F1$_{macro}$  \\
         \citet{AraStance2021:NLP4IF} & ANS & 90.? & F1$_{macro}$   \\
         \citet{AraStance2021:NLP4IF} & AraStance & 78.? & F1$_{macro}$   \\
         \bottomrule
    \end{tabular}
    }
    \caption{
    State-of-the-art results on the stance detection datasets. Note that some papers round their results to integers, and thus we put `\textbf{?}' for them. $^*$Extracted from the FEVER leaderboard.\protect\footnotemark
    }
    \label{tab:sota}
\end{table}

\section{Lessons Learned and Future Trends}
\label{sec:lessons}

\paragraph{Dataset Size} A major limitation holding back the performance of machine learning for stance detection 
is the size of the existing stance datasets, the vast majority of which contain at most a few thousand examples. Contrasted with the related task of Natural Language Inference, where datasets such as SNLI~\citep{bowman-etal-2015-large} of more than half a million samples have been collected, this is far from optimal. {Moreover, the small dataset sizes are often accompanied with skewed class distribution with very few examples from the minority classes, including many of the datasets in this study~\citep{zubiaga-etal-2016-rumor-spread,derczynski-etal-2017-rumoureval,pomerleau-2017-FNC,Baly-2018-IntegratingSD,gorrell-etal-2019-semeval,lillie-etal-2019-joint,AraStance2021:NLP4IF}. This can lead to a significant disparity for label performance (see Section~\ref{sec:methods}). 
Several techniques have been proposed to mitigate this, such as sampling strategies~\citep{Nie2019CombiningFE}, weighting  classes~\cite{veyseh-etal-2017-temporal-attention},\footnote{{Weighting is not trivial for some setups, e.g.,~threaded stance~\citep{ZubiagaKLPLBCA18}}} crafting artificial examples from auxiliary tasks~\cite{yang-etal-2019-blcu,hardalov2022fewshot}, or training on multiple datasets~\citep{schiller2021stance,hardalov2021cross,hardalov2022fewshot}}.

\paragraph{Data Mixing}  A potential way of overcoming limitations in terms of dataset size and focus is to combine multiple datasets. Yet, as we previously discussed (see Section~\ref{sec:factuality}), task definitions and label inventories vary across %
stance  datasets.
Further, large-scale studies of approaches that 
leverage the relationships between label inventories, or the similarity between datasets are still largely lacking. One promising direction is the use of label embeddings~\cite{augenstein-etal-2018-multi}, as they offer a convenient way to learn interactions between disjoint label sets that carry semantic relations. 
One such first study was recently presented by \citet{hardalov2021cross}, which explored different strategies for %
leveraging inter-dataset signals and label interactions in both in- (seen targets) and out-of-domain (unseen targets) settings. 
{This could help to overcome challenges faced by models trained on small-size datasets, and even for smaller minority classes.}%

\paragraph{Multilinguality} 
Multi-linguality is important
for several reasons: (\emph{i})~the content may originate in various languages, (\emph{ii})~the evidence or the stance may not be expressed in the same language, thus (\emph{iii})~posing a challenge for fact-checkers, who might not be speakers of the language the claim was originally made in, and (\emph{iv})~it adds more data that can be leveraged for modelling stance.
Currently, only a handful of datasets for factuality and stance cover languages other than English (see Table~\ref{tab:dataset_features}),
and they are small in size and do not offer a cross-lingual setup. Recently, \citet{vamvas2020xstance} proposed such a setup for three languages for stance in debates, {\citet{schick-schutze-2021-exploiting} explored few-shot learning, and \citet{hardalov2022fewshot} extended that paradigm with sentiment and stance pre-training and evaluated on twelve languages from various domains.} 
{Since cultural norms and expressed linguistic phenomena play a crucial role in understanding the context of a claim~\citep{sap-etal-2019-risk}, we do not argue for a completely language-agnostic framework. 
Yet, empirically, training in cross-lingual setups
improves performance by leveraging better representations 
learned on a similar language 
or by acting as a regulariser.}

\paragraph{Modelling the Context} 
\label{sec:modelling_context}
Modelling the context is a particularly important, yet challenging task. In many cases, there is a need to consider the background of the stance-taker as well as the characteristics of the targeted object. In particular, in the context of social media, one can provide information about the users such as their previous activity, other users they interact most with, the threads they participate in,
or even their interests~\cite{zubiaga-etal-2016-rumor-spread,gorrell-etal-2019-semeval,li-etal-2019-rumor}. The context of the stance expressed in news articles is related to the features of the media outlets, such as source of funding, previously known biases, or credibility~\cite{baly2019multi,darwish2020unsupervised,stefanov-etal-2020-predicting,baly-etal-2020-written}. When using contextual information about the object, factual information about the real world, and the time of posting are all important. Incorporating these into a stance detection pipeline, while challenging, paves the way towards a robust detection process.

\paragraph{Multimodal Content} Spreading mis- and disinformation through multiple modalities is becoming increasingly popular. One such example are \emph{deepfakes}, i.e.,~synthetically created images or videos, in which (usually) the face of one person is replaced with another person's face. Another example are information propagation techniques such as \emph{memetic warfare}. Hence, it is increasingly important to combine different modalities to understand the full context stance is being expressed in.
Some work in this area is on fake news detection for images~\cite{nakamura-etal-2020-fakeddit}, claim verification for images~\cite{zlatkova-etal-2019-fact}, or searching for fact-checked information to alleviate the spread of fake news~\cite{vo-lee-2020-facts}.
There has been work on meme analysis for related tasks: detecting hateful~\cite{Hateful:Memes:Challenge}, harmful \cite{pramanick-etal-2021-detecting,DISARM:2022},  and propagandistic memes~\cite{ACL2021:propaganda:memes,SemEval2021:task6}; see also a recent survey of harmful memes \cite{Survey:2022:Harmful:Memes}.
{This line of research is especially relevant for mis- and disinformation tasks that depend on the wisdom of the crowd in social media as it adds additional information sources~\citep{qazvinian-etal-2011-rumor,zubiaga-etal-2016-rumor-spread,derczynski-etal-2017-rumoureval,hossain-etal-2020-covidlies}; see Section~\ref{sec:modelling_context}}.

\paragraph{Shades of Truth}
The notion of \emph{shades of truth} is important in mis- and disinformation detection. For example, fact-checking often goes beyond binary \emph{true}/\emph{false} labels, e.g.,~\citet{10.1007/978-3-319-98932-7_32} used a third category \emph{half-true},  \citet{rashkin-etal-2017-truth} included \emph{mixed} and \emph{no factual evidence}, and \citet{wang-2017-liar,santia2018buzzface} adopted an even finer-grained schema with six labels, including \emph{barely true} and \emph{utterly false}. We believe that such shades could be applied to stance and used in a larger pipeline. In fact, fine-grained labels are common for the related task of Sentiment Analysis~\cite{pang-lee-2005-seeing,rosenthal-etal-2017-semeval}. 

\paragraph{Label Semantics}
As research in stance detection has evolved, so has the definition of the task and the label inventories, but they still do not capture the strength of the expressed stance. As shown in Section~\ref{sec:factuality} (also Appendix~\ref{sec:stance}, labels can vary based on the use case and the setting they are used in. 
Most researchers have adopted a variant of the \emph{Favour}, \emph{Against}, and \emph{Neither} labels, or an extended schema such as \emph{(S)upport}, \emph{(Q)uery}, \emph{(D)eny}, and \emph{(C)omment}~\cite{mohammad-etal-2016-semeval}, but that is not enough to
accurately assess stance. Moreover, adding label granularity can further improve the transfer between datasets, as the stance labels already share some semantic similarities, but there can be mismatches in the label definitions~\citep{schiller2021stance,hardalov2021cross,hardalov2022fewshot}. %

\paragraph{Explainability} The ability for a model to be able to explain its decisions is getting increasingly important, especially for mis- and disinformation detection, as one could argue that it is a crucial step towards adopting fully automated fact-checking.
The FEVER~2.0 task formulation~\cite{thorne-etal-2019-fever2} can be viewed as a step towards obtaining such explanations, e.g.,~there have been efforts to identify adversarial triggers that offer explanations for the vulnerabilities at the model level~\cite{atanasova-etal-2020-generating-label}.  However, FEVER is artificially created and is limited to Wikipedia, which may not reflect real-world settings. To mitigate this, explanation by professional journalists can be found on fact-checking websites, and can be further combined with stance detection in an automated system. In a step in this direction, ~\citet{atanasova-etal-2020-generating} generated natural language explanations for claims from PolitiFact\footnote{\url{http://www.politifact.com/}} 
given gold evidence document summaries by journalists. 

{Moreover, partial explanations can be obtained automatically from the underlying models, e.g.,~from memory networks~\citep{mitra2018memory}, attention weights~\citep{zhou-etal-2019-gear,liu-etal-2020-fine}, or topic relations~\citep{si-etal-2021-topic}. However, such approaches are limited as they can require gold snippets justifying the document's stance, attention weights can be misleading~\citep{jain-wallace-2019-attention}, and topics might be noisy due to their unsupervised nature.}
Other existing systems~\citep{Popat:2017:TLE:3041021.3055133,Popat:2018:CCL:3184558.3186967,nadeem-etal-2019-fakta} offer explanations to a more limited extent, highlighting span overlaps between the target text and the evidence documents. 
Overall, there is a need for holistic and realistic explanations of how a fact-checking model arrived at its prediction.

\paragraph{Integration} 
People question false information more and tend to confirm true information~\cite{mendoza-etal-2010-twitter-trust}. Thus, stance can play a vital role in verifying dubious content. In Appendix~\ref{sec:appx:applications}, we discuss existing systems and real-world applications of stance for mis- and disinformation identification in more detail.
However, we argue that a tighter integration between stance and fact-checking is needed. Stance can be expressed in different forms, e.g.,~tweets, news articles, user posts, sentences in Wikipedia, and Wiki tables, among others {and can have different formulations as part of the fact-checking pipeline (see Section~\ref{sec:factuality})}. All these can guide human fact-checkers through the process of fact-checking, and can point them to relevant evidence.
Moreover, the wisdom of the crowd can be a powerful instrument in the fight against mis- and disinformation {\cite{Pennycook2521}}, but we should note that vocal minorities can derail public discourse {\cite{doi:10.1080/10810730.2021.1955050}}.
Nevertheless, these risks can be mitigated by taking into account the credibility of the user or of the information source, which can be done automatically or with the help of human fact-checkers.

\section{Conclusion}

We surveyed the current state-of-the-art in stance detection for mis- and disinformation detection. We explored applications of stance for detecting fake news, verifying rumours, identifying misconceptions, and fact-checking. We also discussed existing approaches used in different aspects of the aforementioned tasks, {and we outlined several interesting phenomena, which we summarised as lessons learned and promising future trends}.

\section*{Acknowledgements}

We would like to thank the anonymous reviewers for their useful feedback. 
Isabelle Augenstein's research is partially funded by a DFF Sapere Aude research leader grant with grant number 0171-00034B.
The work is also part of the Tanbih mega-project, which is developed at the Qatar Computing Research Institute, HBKU, and aims to limit the impact of ``fake news,'' propaganda, and media bias by making users aware of what they are reading, thus promoting media literacy and critical thinking.

\bibliography{bibliography}
\bibliographystyle{acl_natbib}

\clearpage
\appendix

\section{Examples of Stance}
\label{sec:appx:examples}

As outlined in Section~\ref{sec:factuality}, there are different formulations in which the task of stance definition is materialised. 
In Table~\ref{tab:examples}, we present some instances of these as exemplified by different stance detection datasets. The target with respect to which the stance is assessed can vary, e.g.,~a headline, a comment, a claim, a topic, etc., which in turn can differ in length and form. Moreover, the context where the stance is expressed can vary not only in its domain, e.g.,~\emph{News} in \cite{ferreira-vlachos-2016-emergent} and \emph{Twitter} in \cite{qazvinian-etal-2011-rumor}, but also in its structure, as seen in the example of multiple evidence sentences in \cite{thorne-etal-2018-fever} and threaded comments in \cite{gorrell-etal-2019-semeval}.

\begin{table*}[t]
    \centering
    \small
    \begin{subtable}[t]{0.49\linewidth}
    \begin{tabularx}{\linewidth}{L}
        \toprule
        \textbf{Headline}: \emph{Robert Plant Ripped up \$800M Led Zeppelin Reunion Contract} \\
        \newspaperLogo \textbf{Body}: {...Led Zeppelin’s Robert Plant turned down £500 MILLION to reform supergroup.. \faThumbsOUp} \\
        \\
        \\
    \end{tabularx}
    \caption{Example from \citet{pomerleau-2017-FNC}}
    \label{tab:examples:fnc}
    \end{subtable}
    \hfill
    \begin{subtable}[t]{0.49\linewidth}
    \begin{tabularx}{\linewidth}{L}
        \toprule
        \textbf{Topic}: \emph{Sarah Palin getting divorced?} \\
        \twitterLogo \textbf{Tweet}: { OneRiot.com - Palin Denies First Dude Divorce Rumors http://url \faThumbsODown} \\
        \hline
        \textbf{Topic}: \emph{N/A (Implicit)} \\
        \twitterLogo \textbf{Tweet}: {Wow, that is fascinating! I hope you never mock our proud Scandi heritage again. \faCommentO}
    \end{tabularx}
    \caption{Examples from \citet{qazvinian-etal-2011-rumor} and \citet{derczynski-etal-2017-rumoureval}}
    \label{tab:examples:twitter}
    \end{subtable}

    \begin{subtable}[t]{0.49\linewidth}
    \begin{tabularx}{\linewidth}{L}
        \toprule
        \textbf{Claim}: \emph{The Rodney King riots took place in the most populous county in the USA.} \\
        \faWikipediaW{iki} \textbf{Evidence 1}: {The 1992 Los Angeles riots, \emph{also known as the Rodney King riots} were a series of riots, lootings, arsons, and civil disturbances that \emph{occurred in Los Angeles County}, California in April and May 1992.} \\
        \faWikipediaW{iki} \textbf{Evidence 2}: {Los Angeles County, officially the County of Los Angeles, \emph{is the most populous county in the USA}. \faThumbsOUp} \\
    \end{tabularx}
    \caption{Example from \citet{thorne-etal-2018-fever}}
    \label{tab:examples:fever}
    \end{subtable}
    \hfill
    \begin{subtable}[t]{0.49\linewidth}
    \begin{tabularx}{\linewidth}{L}
        \toprule
        \textbf{Headline}: \emph{Jess Smith of Chatham, Kent was the smiling sun baby in the Teletubbies TV show} \\
        \newspaperLogo \textbf{Summary 1}: {Canterbury Christ Church University student Jess Smith, from Chatham, starred as Teletubbies sun} \faThumbsOUp \\
        \newspaperLogo \textbf{Summary 2}: {This College Student Claims She Was The Teletubbies Sun Baby} \faThumbsODown
    \end{tabularx}
    \caption{Example from \citet{ferreira-vlachos-2016-emergent}}
    \label{tab:examples:emergent}
    \end{subtable}
    
    \begin{subtable}[t]{1.0\linewidth}
    \begin{tabularx}{\linewidth}{L}
        \toprule
        \makecell[c]{\large \twitterLogo \redditLogo} \\
        \textbf{u1}: We understand that there are two gunmen and up to a dozen hostages inside the cafe under siege at Sydney.. ISIS flags remain on display \#7News \faThumbsOUp \\
        \hspace{2em} \textbf{u2}: @u1 not ISIS flags \faThumbsODown \\
        \hspace{2em} \textbf{u3}: @u1 sorry - how do you know its an ISIS flag? Can you actually confirm that? \faQuestionCircleO \\
        \hspace{4em}\textbf{u4}: @u3 no she cant cos its actually not \faThumbsODown \\
        \hspace{2em}\textbf{u5}: @u1 More on situation at Martin Place in Sydney, AU LINK \faCommentO \\
        \hspace{2em}\textbf{u6}: @u1 Have you actually confirmed its an ISIS flag or are you talking shit \faQuestionCircleO \\
    \end{tabularx}
    \caption{Example from \citet{gorrell-etal-2019-semeval}}
    \label{tab:examples:rumours}
    \end{subtable}
    \caption{Illustrative examples for different stance detection scenarios included in our survey.
    We annotate the expressed stance with \faThumbsOUp\xspace (\emph{support, for}), \faThumbsODown\xspace  (\emph{deny, against}), \faQuestionCircleO\xspace  (\emph{query}), and \faCommentO\xspace  (\emph{comment}).
    }
    \label{tab:examples}
\end{table*}

In a more detailed view of Table~\ref{tab:examples}, we see that each group of examples has its own important specifics that alter the task of stance detection for mis- and disinformation detection.

Figure~\ref{tab:examples:fnc} shows an example from the \emph{News} domain, where we have a headline and an entire article body\footnotetext{For illustrative purposes the text is trimmed to include only the relevant passage.}, and the goal is to find how the two are related in terms of the body's stance(s) towards the headline. In this scenario, the models need to be able to handle very long documents, on one hand, and on the other to reason over multiple fragments of the input text, which might potentially express different stances. It is possible to simplify the task by extracting a summary of the news article beforehand, and evaluating only the stance of that summary, as shown in Figure~\ref{tab:examples:emergent}. However, obtaining such summaries is not a trivial task: (a)~they can be extracted by a human annotator (e.g.,~a journalist), which is time-consuming and expensive, and can require a priori knowledge about the headline/topic of interest as the article might have more than one highlight or viewpoint, or (b)~they can be automatically generated using text summarisation methods, but the result can be noisy.

Stance is often expressed in social media such as Twitter, Facebook, Reddit, etc. We illustrate two such scenarios in Figures~\ref{tab:examples:twitter} and~\ref{tab:examples:rumours}. In contrast to the usually long and well-written news documents, social media posts are mostly short and depend on additional context such as the previous posts in a conversational thread (Figure~\ref{tab:examples:rumours}), or external URLs and implicit topics (Figure~\ref{tab:examples:twitter}). Moreover, these texts also need normalisation, as users tend to use slurs, emojis, and other informal language. 

Next, in Figure~\ref{tab:examples:fever} we highlight another interesting setup: claim verification using multiple pieces of evidence. Here, the reasoning is carried in multiple hops over a set of texts. In particular, there might not exists a single passage from a document/post that supports/refutes the claim directly. In that case, a large enough chain of evidence might be needed, which can cover enough contextual knowledge in order to allow the model (or a person) to assess the veracity of the input claim.

Finally, the examples in Figure~\ref{tab:examples} demonstrate that stance can be used for mis- and disinformation detection in different ways: (\emph{i})~directly, as in the examples in Figures~\ref{tab:examples:fnc} and \ref{tab:examples:twitter}, or (\emph{ii})~as multiple viewpoints, which are later aggregated into a final decision, as in Figure~\ref{tab:examples:fever},~\ref{tab:examples:emergent} and~\ref{tab:examples:rumours}. 

We thoroughly discussed all of the aforementioned setups in Section~\ref{sec:factuality}, including the publicly available datasets that focus on stance in the context of mis- and disinformation identification.

\section{Additional Formulations of Stance as a Component for Fact-Checking}
\label{sec:appx:other_methods}

Beyond the approaches that we outlined in Section~\ref{sec:factuality:component}, stance has also been used for detecting misconceptions and for profiling media sources as part of a fact-checking pipeline. Below, we describe some work that follows these formulations.

\paragraph{Misconceptions} 
\citet{hossain-etal-2020-covidlies} focused on detecting misinformation related to COVID-19, based on known misconceptions listed in Wikipedia.
They evaluated the veracity of a tweet depending on whether it \emph{agrees}, \emph{disagrees}, or has \emph{no stance} with respect to a set of misconceptions. A related formulation of the task is detecting previously fact-checked claims \cite{shaar-etal-2020-known}.
This allows to assess the veracity of dubious content by evaluating the stance of a claim regarding already checked stories, known misconceptions, and facts. 

\paragraph{Media Profiling} 
Stance detection has also been used for media profiling. \citet{stefanov-etal-2020-predicting} explored the feasibility of an unsupervised approach for identifying the political leanings (left, center, or right bias) of media outlets and influential people on Twitter based on their stance on controversial topics. They built clusters of users around core vocal ones based on their behaviour on Twitter such as retweeting, using the procedure proposed by \citet{darwish2020unsupervised}. This is an important step towards understanding media biases.

The reliability of entire news media sources has been automatically rated based on their stance with respect to 
manually fact-checked claims, without access to gold labels for the overall medium-level factuality of reporting~\cite{mukherjee2015leveraging,Popat:2017:TLE:3041021.3055133,Popat:2018:CCL:3184558.3186967}.
The assumption in such methods is that reliable media agree with true claims and disagree with false ones, while for unreliable media, the situation is reversed. %
The trustworthiness of Web sources has also been studied from a data analytics perspective, e.g.,~\citet{Dong:2015:KTE:2777598.2777603} 
proposed that a trustworthy source is one that contains very few false claims.

More recently, \citet{baly2018predicting} used gold labels from Media Bias/Fact Check,\footnote{\url{http://mediabiasfactcheck.com}}
and a variety of information sources: articles published by the medium, what is said about the medium on Wikipedia, metadata from its Twitter profile, URL structure, and traffic information. In follow-up work, \citet{baly2019multi} used the same representation to jointly predict a medium's factuality of reporting (\emph{high} vs. \emph{mixed} vs. \emph{low}) and its bias (\emph{left} vs. \emph{center} vs. \emph{right}) on an ordinal scale, in a multi-task ordinal regression setup.

\citet{baly-etal-2020-written} extended the information sources to include Facebook followers and speech signals from the news medium's channel on YouTube. 
Finally, \citet{hounsel2020identifying} proposed to use domain, certificate, and hosting information about the infrastructure of the website. See \cite{Survey:2021:Media:Factuality:Bias} for a recent survey on media profiling.

There is a well-known connection between factuality and bias.\footnote{\url{http://www.poynter.org/fact-checking/media-literacy/2021/should-you-trust-media-bias-charts/}}
For example, hyper-partisanship is often linked to low trustworthiness~\cite{Potthast2018}, e.g., appealing to emotions rather than sticking to the facts, while center media tend to be generally more impartial and also more trustworthy. 

\paragraph{User Profiling}
In the case of social media and community fora, it is important to model the trustworthiness of the user. In particular, there has been research on finding opinion manipulation \emph{trolls}, paid \cite{Mihaylov2015ExposingPO} 
or just perceived \cite{Mihaylov2015FindingOM},
\emph{sockpuppets} \cite{Maity:2017:DSS:3022198.3026360,Kumar:2017:AMS:3038912.3052677}, \emph{Internet water army} \cite{Chen:2013:BIW:2492517.2492637}, and \emph{seminar users} \cite{SeminarUsers2017}.

\section{Systems and Applications}
\label{sec:appx:applications}

The systems and applications below use stance detection as part of a pipeline for identifying mis- and disinformation, see Section~\ref{sec:methods} for more details about the methods.

\citet{wen-etal-2018-cross} worked in a cross-lingual cross-platform rumour verification setup. They included multimodal content from fake and from real posts with images or videos shared on Twitter. They then collected supporting documents from two search engines,~Google and Baidu, in English and Chinese, which they used for veracity evaluation. They trained their stance detection model on English data (FNC-1) using pre-trained multilingual sentence embeddings, and further added cross-platform features in their final neural model. 

\citet{Popat:2018:CCL:3184558.3186967} proposed CredEye,\footnote{\url{https://gate.d5.mpi-inf.mpg.de/credeye/}} 
a system for automatic credibility assessment of textual claims. The system takes a claim as an input and analyses its credibility by considering relevant articles it retrieved from the Web, by combining the predicted stance of the articles regarding the claim with linguistic features to obtain a credibility score~\citep{Popat:2017:TLE:3041021.3055133}.

\citet{DBLP:conf/aaai/NguyenKLW18} designed a prototype fact-checker Web tool.\footnote{\url{http://fcweb.pythonanywhere.com/}} Their system leverages a probabilistic graphical model to assess a claim's veracity taking into consideration the stance of multiple articles regarding this claim, the reputation of the news sources, and the annotators' reliability. In addition, it offers explanations to the fact-checkers based on the aforementioned features, which was shown to improve the overall user satisfaction and trust in the predictions.

\citet{zubiaga-et-al-2018-socialmedia-survey} considered a four-step tracking process as a pipeline for rumour verificatioon: (\emph{i})~\emph{rumour detection}, i.e.,~given a stream of claims, determine whether they are worth verifying or they do contain no rumours, (\emph{ii})~\emph{rumour tracking} for finding relevant information about the rumour using social media posts, sentence descriptions, and keywords, (\emph{iii})~\emph{stance classification} to collect stances towards the rumour, and (\emph{iv})~\emph{veracity classification} to aggregate the information from the tracking component, the collected stances, and optionally other relevant information about the sources, metadata about the users, etc., to predict a truth value for the rumour.

\citet{nadeem-etal-2019-fakta} developed FAKTA, a system for automatic end-to-end fact-checking of claims. It retrieves relevant articles from Wikipedia and as well as from selected media sources, which it then uses for verification. FAKTA uses a stance detection model, trained in a FEVER setting, to predict the stance and to obtain entailed spans. These predictions, combined with linguistic analysis, are used to provide both document- and sentence-level explanations and a factuality score.

\citet{nguyen-et-al-FANG} proposed the Factual News Graph (FANG) model, which models the social context for fake news detection. In particular, FANG uses the stance of user comments with respect to the target news article as an integral component of its model, together with temporality, user--user interactions, article--source interactions, as well a source reliability information.

\end{document}